# Y$\mathcal{GG}$DRASIL – A statistical package for learning Split Models

Søren Højsgaard *
Biometry Research Unit
Danish Institute of Agricultural Sciences
Denmark

## Abstract

There are two main objectives of this paper. The first is to present a statistical framework for models with context specific independence structures, i.e. conditional independencies holding only for specific values of the conditioning variables. This framework is constituted by the class of split models. Split models are an extension of graphical models for contingency tables and allow for a more sophisticated modelling than graphical models. The treatment of split models include estimation, representation and a Markov property for reading off those independencies holding in a specific context. The second objective is to present a software package named YGGDRASIL which is designed for statistical inference in split models, i.e. for learning such models on the basis of data.

## 1 INTRODUCTION

Recently there has been an increased interest in models which explicitly account for conditional independencies holding only for specific values of the variables conditioned upon. This phenomenon is here referred to as context specific independence.

With such models, a more sophisticated and realistic modelling can be achieved compared with graphical models, where focus is on conditional independence restrictions. See Lauritzen (1996) for comprehensive treatment of graphical models. For comparison, one can think of conditional independence as a context specific independence which holds for all values of the conditioning variables. Similar ideas are discussed in connection with Bayesian networks by e.g. Geiger and Heckerman (1991), Boutilier, Friedman, Goldszmidt and Koller (1996).

There are two main objectives of this paper. The first is to introduce the class split models, which is an extension of (undirected) graphical models for contingency tables. In split models, the fundamental property of interest is context specific independence. Split models are graphical in the sense that any split model admits a graphical representation – not by a single graph but by a collection of successively simpler graphs arranged in a hierarchical structure. Split models are described in Section 3

The second objective is to present a software package named Y$\mathcal{GG}$DRASIL designed for statistical inference in split models. That is, for learning such models on the basis of a multidimensional contingency table. The program is described in Section 4.

Prior to treating the two main objectives, a general class of models denoted Context Specific Interaction Models (hereafter abbreviated CSI models) is described. This is done in Section 2. CSI models can be regarded as a special class of log–linear models for contingency tables and all split models are CSI models. The exposition of CSI models includes results regarding estimation and a Markov property by which conditional and context specific independencies implied by a CSI model can be read off from an undirected graph.

## 2 CONTEXT SPECIFIC INTERACTION MODELS

### 2.1 NOTATION

Let $R$ denote a set of random variables where each $r \in R$ takes values in a finite state space $\mathcal{I}_r$. Hence a configuration of $R$ takes values in the product space $\mathcal{I}_R = \times_{r \in R} \mathcal{I}_r$. A typical element of $\mathcal{I}_R$ is denoted by $i$. A subset $A \subset R$ takes values in $\mathcal{I}_A = \times_{r \in A} \mathcal{I}_r$, where $i_A$ denotes a typical element.

---

*Biometry Research Unit, Danish Institute of Agricultural Sciences, Research Centre Foulum, Blichers Allé, DK-8830 Tjele, Denmark.

Associated with $R$ there is a contingency table and the number of objects in cell $i$ is $n(i)$. Correspondingly, the marginal table given by $A$ has $n(i_A)$ objects in cell $i_A$. For a fixed configuration $i_A^*$ we obtain a slice of the table with counts $n(i_A^*, i_{R \setminus A})$, for $i_{R \setminus A} \in \mathcal{I}_{R \setminus A}$. Interest is in cell probabilities $p(i)$ of an object falling in cell $i$, in the corresponding marginal probabilities $p(i_A)$ obtained by summing $p(i)$ over $R \setminus A$, and in the $i_A^*$-slice of $p$, i.e. $p(i_A^*, i_{R \setminus A})$.

## 2.2 CONTEXT SPECIFIC INDEPENDENCE

Let $A$, $B$, $S$ and $E$ denote disjoint sets of random variables and let $i_E^*$ denote a particular value of $E$. If $p(i_A, i_B | i_S, i_E^*) = p(i_A | i_S, i_E^*) p(i_B | i_S, i_E^*)$ for all values of $S$ such that $p(i_S, i_E^*) > 0$, it is said that $A$ and $B$ are *context specific independent* given $S$ and $E = i_E^*$, written $A \perp\!\!\!\perp B | (i_E^*, S)$. This is equivalent to a factorization of the $i_E^*$-slice of $p$ as

$$p(i_A, i_B, i_S, i_E^*) = g(i_A, i_S) h(i_B, i_S), \quad (1)$$

where $g$ and $h$ are non-negative functions.

Hence if $A \perp\!\!\!\perp B | (i_E^*, S)$ for all values of $i_E^* \in \mathcal{I}_E$ then $A$ and $B$ are *conditionally independent* given $(S, E)$, written $A \perp\!\!\!\perp B | (S, E)$. Thus conditional independence is a special case of context specific independence in the sense that the context specific independence holds in all contexts.

## 2.3 CSI MODELS

For a subset $A \subset R$, a potential $u_A$ is a non-negative real valued function defined on $\mathcal{I}_R$ which depends on $i$ only through $i_A$. For disjoint subsets $A$ and $b$ of $R$ and a particular (fixed) configuration $j_b \in \mathcal{I}_b$ of $b$ we define a context potential as $u_{A \cup b}(i_A; i_b)^{1_{\{i_b = j_b\}}}$ where $1_{\{\}}$ is an indicator function. We use the notation $u_A^{j_b}(i_A; i_b)$ for such a term and note that $u_A^{j_b}$ is constantly equal to one for all $i \in \mathcal{I}_R$ for which $i_b \neq j_b$.

A pair $(A, j_b)$ as given above is denoted a generator (and is sometimes also written $A^{j_b}$). A set $\mathcal{C}$ of generators is denoted a generating class. A generating class $\mathcal{C}$ defines the model through the model function

$$p(i) = \prod_{A^{j_b} \in \mathcal{C}} u_A^{j_b}(i_A; i_b). \quad (2)$$

Formally a CSI model $\mathcal{M}(\mathcal{C})$ with generating class $\mathcal{C}$ is the set of probabilities with the form in (2).

## 2.4 ESTIMATION IN CSI MODELS

A CSI model $\mathcal{M}(\mathcal{C})$ is a hierarchical log-affine model in the sense of (Lauritzen 1996), Chapter 4. Hence, under multinomial sampling, the maximum likelihood estimate for $p$ is given as the unique solution to the system of equations

$$np(i_A, j_b) = n(i_A, j_b) \quad (3)$$

for all $i_A \in \mathcal{I}_A$ for all $A^{j_b} \in \mathcal{C}$. An iterative proportional scaling procedure (guaranteed to converge) for solving (3) can be found in Højsgaard (2000). This algorithm is implemented in Y$\mathcal{GG}$DRASIL. It is can be noted that a CSI model is not in general constituted by different log-linear models applied separately to different slices of the table.

## 2.5 INTERPRETATION OF CSI MODELS

The key to interpreting CSI models is a graphical representation of the interaction structure holding in any given context. Before establishing such a representation it is noted that any hierarchical model can be represented by its interaction graph. This is the graph whose vertices correspond to the variables and whose edges correspond to the 2-factor interactions in the model. (For example, if the model contains a 3-factor interaction $XYZ$ the model also contains the two factor interactions $XY$, $XZ$, and $YZ$). From the interaction graph all conditional independencies implied by the model can be read off using the rule that if $D$ separates $A$ and $B$ in the interaction graph, then $A \perp\!\!\!\perp B | D$. This rule is for graphical models known as the *global Markov property* and provides an easy tool for reading off conditional independencies entailed by the model.

In connection with CSI models interest is in general in the structural form of $p(i_{R \setminus E} | i_E^*)$, where $i_E^*$ is a specific configuration of the variables in $E \subset R$. When conditioning on $i_E^*$, attention is restricted to the sample points $\mathcal{I}^* = \{i \in \mathcal{I}_R | i_E = i_E^*\}$, i.e. the $i_E^*$-slice of $p$ where $p$ is given by (2).

For a fixed configuration $i_E^*$ we define $\mathcal{C}(i_E^*)$ as those sets $A \cup b$ for which the corresponding generators $A^{j_b} \in \mathcal{C}$ match with the configuration $i_E^*$, i.e.

$$\mathcal{C}(i_E^*) = \{A \cup b | (A, j_b) \in \mathcal{C} \text{ and } j_{b \cap E} = i_{b \cap E}^*\}.$$

The rationale is that potentials corresponding to generators which *do not* match (i.e. generators $A^{j_b}$ for which $j_{b \cap E} \neq i_{b \cap E}^*$) are constantly equal to one on $\mathcal{I}^*$. The graph generated by $\mathcal{C}^*(i_E^*)$ is said to be the graph *instantiated* by $i_E^*$ and is denoted $G_{inst}(i_E^*)$.

Hence when restricted to $\mathcal{I}^*$, $p$ has interactions only among variables in $\mathcal{C}(i_E^*)$, i.e. only among neighbours in $G_{inst}(i_E^*)$. The fact that if $X \perp\!\!\!\perp Y|Z$ and if $U = h(X)$ then $U \perp\!\!\!\perp Y|Z$ gives when applied to (1) what shall be referred to as the *global CSI Markov property*:

**Theorem 1** *Let $G_{inst}(i_E^*)$ be the graph instantiated by $i_E^*$. Then $A \perp\!\!\!\perp B|(S, i_E^*)$ whenever $S \cup E$ separates $A$ and $B$ in $G_{inst}(i_E^*)$*

This result provides an easy tool for reading off context specific and conditional independencies emerging when conditioning on particular values of particular variables.[1] Thus for example $p(i_R)$ is represented by the graph generated by $\{A \cup b|(A, j_b) \in \mathcal{C}\}$.

In Y$\mathcal{GG}$DRASIL various facilities are available for generating instantiated graphs for arbitrary CSI models.

## 3 SPLIT MODELS

A split model is a CSI model whose generating class is given by a split graph which is a collection of successively simpler graphs arranged in a hierarchical structure. The components of a split graph are introduced in the following through examples. For the formal definitions we refer to the Y$\mathcal{GG}$DRASIL documentation, see Section 4 for references.

It is illustrative to think of the process described below as follows: Suppose that on the basis of data and some model selection scheme a graphical model $G$ has been found representing the conditional independence restrictions holding in the domain. Then focus is turned to revealing more details by searching for additional context specific independencies.

### 3.1 SPLIT TREES

In the following examples we shall for simplicity assume that all variables are binary and that a variable $A$ takes values in $\{a^+, a^-\} \equiv \{+, -\}$. We shall also identify a graph $G$ with its cliques $\mathcal{C}(G)$ such that the graph $G_U$ in Figure 1, (1) is written $[ABD][ACD]$, while the graphs $G_1$ and $G_2$ in Figure 1, (2) are both $[BD][CD]$. (The reason for the subscript $U$ becomes apparent below).

The tree in Figure 1, (2) is denoted a *split tree* and the pair $(a^+, G_1)$ consisting of the graph $G_1$ together with the value $a^+$ of $A$ with which it is associated is called a *context graph*. This is written as $G_1^{a^+}$ in short and the split tree is written $\mathcal{ST}_U = (G_1^{a^+}, G_2^{a^+})$.

It is then said that a *split* is made by $A$ in the graph $G_U$. A context graph identifies a generating class which for $G_1^{a^+}$ is $\mathcal{C}(G_1^{a^+}) = \mathcal{C}(G_1) \times \{a^+\}$ (written shortly as $[BD, a^+][CD, a^+]$ or $[[BD][CD]]^{a^+}$). Correspondingly $\mathcal{ST}_U$ identifies the generating class $\mathcal{C}(\mathcal{ST}_U) = [[BD][CD]]^{a^+}[[BD][CD]]^{a^-}$. It is clear that $\mathcal{ST}$ and $G$ specify the same model. Thus a split in itself represents here no change of the model.

---
[1]Note that since the variables in $E$ are kept fixed, i.e. conditioned on, these variables can be eliminated from $G_{inst}(i_E^*)$ if so desired.

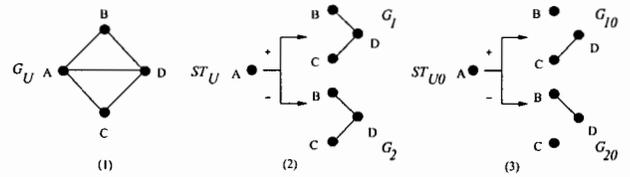

Figure 1: (1) Undirected graph, (2) equivalent split tree, (3) split tree obtained after elimination of context edges.

An edge in a context graph is called a *context edge* and the graph $G_{10}$ in Figure 1, (3) is obtained by removing the context edge $(\{B, D\}, a^+)$ (written shortly as $\{B, D\}^{a^+}$) from $G_1^{a^+}$. The split tree $\mathcal{ST}_{U0}$ in Figure 1, (3) specifies the generating class $\mathcal{C}(\mathcal{ST}_{U0}) = [[B][CD]]^{a^+}[[BD][C]]^{a^-}$ which is a reduction of the model given by $\mathcal{ST}_U$. The model given by $\mathcal{ST}_{U0}$ satisfies the constraints implied by $G_U$, i.e. $C \perp\!\!\!\perp B|(A, D)$, but also the additional restrictions $B \perp\!\!\!\perp (C, D)|a^+$ and $C \perp\!\!\!\perp (B, D)|a^-$. This can be seen directly or by forming the relevant instantiated graphs.

### 3.2 SPLIT GRAPHS

Next consider the graph $G$ in Figure 2, (1) with cliques $[ABD][ACD][ABE][AF][CG][FG]$. Letting $U = \{ABCD\}$ it follows that $G_U$ in Figure 1 is the subgraph of $G$ induced by $U$.

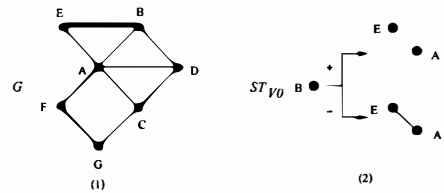

Figure 2: (1) Undirected graph, (2) split tree. The pairs $\mathcal{SG} = (G, \{\mathcal{ST}_{U0}\})$ and $\mathcal{SG}^* = (G, \{\mathcal{ST}_{U0}, \mathcal{ST}_{V0}\})$ are split graphs.

The operation of reducing $G_U$ to $\mathcal{ST}_{U0}$ can also be embedded as an operation on the larger graph $G$ as follows: First a collection of cliques, here $\mathcal{H}_1 = \{[ABD][ACD]\}$ with variables $U = \{ABCD\}$ is chosen and hereby also a corresponding subgraph $G_U$ of $G$. Next certain reductions of $G_U$ are made giving

$\mathcal{ST}_{U0}$. The pair

$$SG = (G, \{\mathcal{ST}_{U0}\}) \qquad (4)$$

is said to be a *split graph*. This split graph identifies a generating class $\mathcal{C}(SG)$ for a CSI model which is achieved by 1) removing $\mathcal{H}_1$ from $\mathcal{C}(G)$ and 2) adding the generating class specified by $\mathcal{ST}_{U0}$. That is,

$$\begin{aligned}\mathcal{C}(SG) &= \mathcal{C}(\mathcal{ST}_{U0}) \cup (\mathcal{C}(G) \setminus \mathcal{H}_1) \\ &\equiv [CD]^{a^+}[BD]^{a^-}[ABE][AF][CG][FG],\end{aligned} \qquad (5)$$

where redundant terms have been removed in (5).

Additional splits can be made in $G$. For instance we can pick $\mathcal{H}_2 = \{[ABE]\}$ with variables $V = \{ABE\}$, split by $B$ and eliminate the context edge $\{A, E\}^{b^+}$ as illustrated in $\mathcal{ST}_{V0}$ in Figure 2, (2). After doing so the split graph becomes

$$SG^* = (G, \{\mathcal{ST}_{U0}, \mathcal{ST}_{V0}\}).$$

The corresponding generating class $\mathcal{C}(SG^*)$ becomes, after eliminating redundant terms,

$$[CD]^{a^+}[BD]^{a^-}[B][E][AE]^{b^+}[AF][CG][FG].$$

### 3.3 THE GENERAL CASE

In the previous section, each split tree was a list of context graphs, where a context graph was a pair consisting of a context and an undirected graph. In turn, a split graph was constituted by an undirected graph combined with a set of split trees.

One can imagine also making splits in a context graph such that a context graph is turned into a split graph. To facilitate this case, the general definition of split trees and split graphs is recursive: A split tree is a list of split graphs. A split graph in turn is a triple consisting of a (possibly empty) context, an undirected graph and a (possibly empty) list of split trees. Hence a split graph with an empty list of split trees is simply a context graph and if the context is also empty, the context graph is simply a graph. This recursive definition makes sense because after a split we are always left with graphs with fewer vertices than the graph we made the split in.

### 3.4 TWO CAVEATS

There are some caveats in connection with making splits which can be illustrated on the basis of the split graph $SG$.

Suppose that in $[ABE]$ a split is instead made by $E$ followed by removal of the context edge $\{A, B\}^{e^+}$ leading to the split tree $\mathcal{ST}'_U = (\{[A][B]\}^{e^+}, \{[AB]\}^{e^-})$. Removal of $\{A, B\}^{e^+}$ does not entail any context specific independence because the $[ABD]$ interaction term is present in $\mathcal{C}(SG)$. Hence such a split has no meaning.

Another caveat is that one could be tempted to split by say $C$ in the graph $G_W$ induced by $W = \{ACFG\}$. This would lead to the split tree $\mathcal{ST}_W = ([[AF][FG]]^{c^+}, [[AF][FG]]^{c^-})$ which specifies the same model as $[ACF][CFG]$ does. Under $G$ it holds that $F \perp\!\!\!\perp C|(A, G)$ but this conditional independence is not implied by the model specified by $\mathcal{ST}_W$. Hence, this latter split would change the model as interactions are to the original model. But then one can no longer take $G$ to represent the overall structure of association holding in all contexts.

Y$\mathcal{GG}$DRASIL ensures that no such illegal or meaningless splits are made.

### 3.5 DECOMPOSITIONS

For brevity the model specified by $G$ is written $\mathcal{M}(G)$ rather than $\mathcal{M}(\mathcal{C}(G))$ and likewise for split trees and split graphs.

The model $\mathcal{M}(G)$ can be decomposed into the models $\mathcal{M}_1 = \mathcal{M}([ABD][ACD])$, $\mathcal{M}_2 = \mathcal{M}([ABE])$, and $\mathcal{M}_3 = ([AC][AF][CG][FG])$. Here $\mathcal{M}_1$ is a marginal model for $ABCD$ and estimation in this model can be based on the $ABCD$-marginal table only.

From (3) it follows that a $\mathcal{M}(SG)$ admits a corresponding decomposition into $\mathcal{M}(\mathcal{ST}_{U0})$, $\mathcal{M}_2 = \mathcal{M}([ABE])$, and $\mathcal{M}_3 = ([AC][AF][CG][FG])$. See Højsgaard (2000) for details. Thus testing $\mathcal{M}(SG)$ under $\mathcal{M}(G)$ corresponds to testing $\mathcal{M}(\mathcal{ST}_{U0})$ under $\mathcal{M}(\mathcal{ST}_U)$, where the latter model is equal to $\mathcal{M}(G_U)$.

In this latter test another type of decomposition comes into play. The model given by $\mathcal{ST}_{U0}$ decomposes into a model for the $a^+$–context given by $\mathcal{C}(G_{10})$ and a model for $a^-$–context given by $\mathcal{C}(G_{20})$. These two models are completely unrelated and the test for $\mathcal{M}(\mathcal{ST}_{U0})$ under $\mathcal{M}(G_U)$ factorizes into two independent tests: One for testing $B \perp\!\!\!\perp C|a^+$ which is based on the $a^+$–slice of the $ABCD$–marginal table and a corresponding test for $D \perp\!\!\!\perp C|a^-$ in the $a^-$–slice of the $ABCD$–marginal table.

In Y$\mathcal{GG}$DRASIL such decompositions are heavily exploited in connection with model selection and estimation.

## 4 Y$\mathcal{GG}$DRASIL

The remaining part of this paper is dedicated to illustrating the functionality of Y$\mathcal{GG}$DRASIL. Information about Y$\mathcal{GG}$DRASIL (including a more comprehensive documentation of the theory and a users guide) can be found at the Y$\mathcal{GG}$DRASIL homepage on

```
http://www.jbs.agrsci.dk/~sorenh/yggdrasil.html
```

Y$\mathcal{GG}$DRASIL facilitates estimation and test in CSI model including split models. Facilities for obtaining parameter estimates, fitted values etc. are provided. Also an automated procedure for model search in split models is available. Y$\mathcal{GG}$DRASIL is an extension of Xlisp+CoCo (Badsberg 1995), which refers to the functionality of the program CoCo, (Badsberg 1991) loaded into the Lisp dialect Xlisp-Stat (Tierney 1990). CoCo is a program for estimation, test and model search among hierarchical interaction models for large contingency tables. Xlisp-Stat is an object oriented environment for statistical computing. Thereby Xlisp+CoCo provides the user with unique facilities for handling hierarchical (and thereby especially graphical) interaction models in an object oriented statistical programming environment.

### 4.1 WOMEN AND MATHEMATICS

Fowlkes, Freeny and Landwehr (1988) report a survey among 1,190 New Jersey high school students in connection with a campaign for encouraging interest in mathematics especially among females. A part of the campaign was a series of lectures by women working in the mathematical sciences.

Eight schools (four urban and four suburban) were included in the study, and the students were allegedly assigned for attendance or non–attendance in the lectures. Data on six binary variables were collected: *Attendance in math lectures* (A) (attended=1, did not attend=2), *Sex* (B) (female=1, male=2), *School type* (C) (suburban=1, urban=2), *Agree in statement "I'll need mathematics in my future work"* (D) (agree=1, disagree=2), *Subject preference* (E) (math–science=1, liberal arts=2), *Future plans* (F) (college=1, job=2). We shall often refer to the variable D as *Attitude* (towards mathematics) and part of the study was to investigate whether *Attitude* (D) and *Subject preference* (F) are related to the lecture attendance. Data were also analyzed in Upton (1991), where the data can be found. The variables $A$, $B$, and $C$ can be regarded as explanatory and we therefore only consider models containing the $ABC$ 3–factor interaction.

### 4.2 GETTING STARTED

The contingency table is put into the list **wam-data**. The next step is to create a coco–object, here referred to as **wam**, which can be regarded as a session with the stand alone version of CoCo. Subsequently a specification of data and data itself is entered into this object. In the examples found below, commands given by the user to the system are preceded by >:

```
>(def wam-data '(
37 27  51 48 16 11 10 19 16 15  7  6 12 24 13  7
10  8  12 15  9  4  8  9  7 10  7  3  8  4  6  4
51 55 109 86 24 28 21 25 32 34 30 31 55 39 26 19
 2  1   9  5  8  9  4  5  5  2  1  3 10  9  3  6))
>(def wam (make-coco))
>(send wam :enter-names
      '( "A" "B" "D" "E" "F" "C")
      '(  2   2   2   2   2   2 ))
>(send wam :enter-table wam-data)
```

Following these initial specifications the road is paved for a statistical analysis. First it is specified the $ABC$ 3–factor interaction has to be contained in any model. This is achieved by sending the message :fix-edges to the **wam** object. Subsequently we create the saturated model which is represented by a graph object here denoted G-sat:

```
>(send wam :fix-edges "ABC")
>(def G-sat (send wam :make-graph :model "ABCDEF"))
```

We proceed by selecting a graphical model by applying a stepwise backward elimination procedure on 5% significance level starting from the saturated model. This is achieved by sending the message :drop-least with some additional specifications to the graph object G-sat. This results in creation of a new graph object which we name G:

```
>(def G (send G-sat :drop-least :recursive T
              :decomposable-mode T :p-accepted 0.05))
```

The graph G is shown in Figure 3. The generating class of the model and some summary statistics can be achieved by:

```
>(send G :summary)

Counts Deviance  df p-value   AIC Model
  1190   23.284  32 0.86914 -40.72 [[ABEC][BDEC][DEFC]]
```

The deviance is defined as minus twice the log likelihood ratio test statistic between the model under consideration and the saturated model. The AIC statistic is Akaikes Information Criterion, (Akaike 1974) and is given as the deviance minus twice the degrees of freedom. Small values of AIC indicates a good fit of the model.

### 4.3 PARTITIONING TESTS

The 4–factor interaction $[BCDE]$ in G implies an association between *Sex* (B) and *Attitude* (D) in the sense that $B$ and $D$ are not conditionally independent given some other variables. The test for deletion of the $\{B, D\}$-edge is rejected (otherwise the edge would have been removed in the stepwise procedure above).

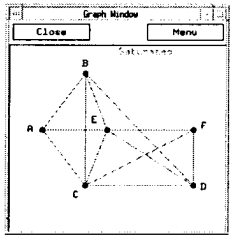

Figure 3: Graph G for the selected graphical model $[ABCE][BCDE][CDEF]$.

The test for deletion of $\{B, D\}$ can however be partitioned in various ways. One such (not shown) is as 4 separate tests for independence between $B$ and $D$ in the slices defined by the values of $C$ and $E$. Another partitioning is as two test for $D \perp\!\!\!\perp B|E, C = c$ – one for each value $c$ of $C$, and two similar tests for each value of $E$:

```
>(send G :split-test-edge "BD" :partition 'single)
```

```
Counts  Deviance  df  p-value   AIC     C
  443     1.851    2   0.39639  -2.15   (1)
  747    23.657    2   0.00001  19.66   (2)
---------------------------------------------
 1190    25.507    4   0.00004  17.51   Total

Counts  Deviance  df  p-value   AIC     E
  736    17.983    2   0.00012  13.98   (1)
  454     7.525    2   0.02323   3.52   (2)
---------------------------------------------
 1190    25.507    4   0.00004  17.51   Total
```

From the partitioning by $C$ above it follows that the main contribution to the overall test statistic on 25.507 is due to the $C = 2$–slice, i.e. to the urban schools. Note that the AIC statistic is additive over the levels of $C$ and $E$. The negative value for the $C = 1$–slice indicates like the significance test independence between *Attitude* and *Sex* in suburban schools when adjusting for *Subject preference*. Hence the results suggest a reduction of the 4–factor interaction $[BCDE]$ to $[[BE][DE]]^{C=1}[[BDE]]^{C=2}$.

It can be argued that one should consider an adjustment of the significance level when partitioning such a test, just as one should do in connection with stepwise model selection in for instance graphical models. However, to our knowledge no satisfactory scheme for doing so exists.

### 4.4 SELECTING A SPLIT MODEL

The results above indicate that more structure can be revealed by looking for context specific independencies. However, it is also clear that a systematic approach for looking for context specific independencies is needed, and this is what split models provide.

In Y$\mathcal{GG}$DRASIL automated approaches for selecting split models are available. One such is to regard the model as being constituted by atoms, in general its irreducible components which in the decomposable case are the cliques of the model. In each clique one then aims for the split giving the best split tree. Since a split in itself does not imply any model reduction, a split has to be followed by a model selection scheme in each context graph. In this connection the selection scheme leading to the graph G above was applied. The created split graph is named SG, and we ask for having the generating class $\mathcal{C}(\text{SG})$ returned:

```
>(def SG (send G :split-drop-least :recursive T
              :p-accepted 0.05))
>(send SG :return-model)

(("C" (1) "[[ABE]]") ("C" (2) "[[AB][BE]]")
 ("C" (1) "[[BE][DE]]") ("C" (2) "[[BDE]]")
 ("D" (1) "[[EFC]]") ("D" (2) "[[EC][FC]]"))
```

Hence the optimal split in the cliques $[ABCE]$ and $[BCDE]$ is by $C$ (*School type*), while in $[CDEF]$ it is by $D$ (*Attitude*).

The interpretation of the latter split is that $F \perp\!\!\!\perp E|C, D = 2$ where $D = 2$, i.e. *Future plans* is independent of *Subject preference* given *School type* when $D = 2$, that is for students who do not believe that they will need mathematics in their future work.

Rather than treating $[ABCE]$ and $[BCDE]$ separately one could decide to treat these two components simultaneously aiming for a common split. The advantage of this is there are fewer restrictions of the kind illustrated in Section 3.4 to comply with. For instance, in connection with a split $E$, the $\{B, C\}$-edge is now eligible for removal. That was not the case in forming SG above. The price to pay is that fewer variables are eligible as possible split variables, in fact only $B$, $C$, and $E$. This is accomplished as follows:

```
>(def SG2 (send G :split-drop-least
    :recursive T :p-accepted 0.05
    :collection '(("[BDEC]" "[ABEC]") "[DEFC]")))
>(send SG2 :return-model)

(("C" (1) "[[ABE][DE]]") ("C" (2) "[[AB][BDE]]")
 ("D" (1) "[[EFC]]") ("D" (2) "[[EC][FC]]"))
```

As can be seen, no additional model reduction was achieved in SG2 compared with SG. That is, the $\{B, E\}$-edge was significant for both levels of $C$. The split graph SG2 is formally given by

$$\begin{aligned}
\text{SG2} &= (\text{G}, \{\mathcal{ST}_{ABCDE}, \mathcal{ST}_{CDEF}\}), \text{ where} \\
\mathcal{ST}_{ABCDE} &= ([[ABE][DE]]^{C=1}; [[AB][BDE]]^{C=2}) \\
\mathcal{ST}_{CDEF} &= ([[EFC]]^{D=1}; [[EC][FC]]^{D=2}),
\end{aligned}$$

and is shown in Figure 4. In this example, a split is made in every component of the graph G and the graph itself therefore contains no information about the generating class for the corresponding model.

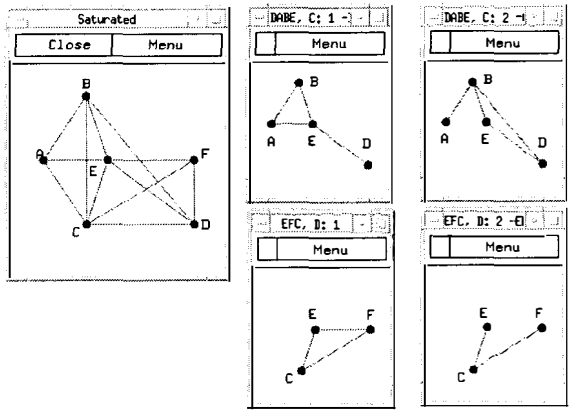

Figure 4: The split graph SG2 selected on the basis of G.

Summary statistics for SG2 are obtained as:

```
>(send SG2 :summary)

Counts Deviance df p-value   AIC    Model
   443    1.851  2 0.39639 -2.15  G(C=1): [[DE][ABE]]
   747    2.291  2 0.31800 -1.71  G(C=2): [[BDE][AB]]
-----------------------------------------------
  1190    4.142  4 0.38711 -3.86  ST()[CDABE]

Counts Deviance df p-value   AIC    Model
   735    0.000  0 0.00000  0.00  G(D=1): [[EFC]]
   455    1.671  2 0.43372 -2.33  G(D=2): [[EC][FC]]
-----------------------------------------------
  1190    1.671  2 0.43372 -2.33  ST()[DEFC]

Counts Deviance df p-value   AIC    Model
  1190   23.284 32 0.86914 -40.72 [[ABEC][BDEC][DEFC]]
  1190    4.142  4 0.38711  -3.86 ST()[CDABE]
  1190    1.671  2 0.43372  -2.33 ST()[DEFC]
-----------------------------------------------
  1190   29.097 38 0.84987 -46.90 SG()[ABDEFC]
```

The output from Y$\mathcal{GG}$DRASIL is to be read bottom-up: The total deviance on 29.097 is partitioned into the deviance on 23.284 for the graph G plus the deviance contributions from the two split trees in SG2. The deviance for each split tree can be found above and is partitioned according to the levels of the split variable.

Since $D$ is a response it can be argued that the latter split does not make much sense. Hence one may a priori decide to exclude a split by $D$ (and $F$). Such constraints on the model selection scheme can be imposed in Y$\mathcal{GG}$DRASIL. We abstain from illustrating this.

In the example above, only one level of splits were made, i.e. there were made splits in the graph G only. As indicated in Section 3.3 the splitting process can go on recursively by also making splits in context graphs. Such an approach to model selection is also available in Y$\mathcal{GG}$DRASIL.

### 4.5 CREATING INSTANTIATED GRAPHS

The split graph is believed to be of most use in connection with the inference part. When it comes to interpretation in terms of context specific independencies, the instantiated graphs are more useful.

It is clear that G itself is the interaction graph for the model specified by SG2, and hence the graph instantiated when not conditioning on anything. The instantiated graphs corresponding to the levels of $C$ can be created by:

```
>(send SG2 :instantiate-all "C")
```

The resulting graphs are shown in Figure 5.

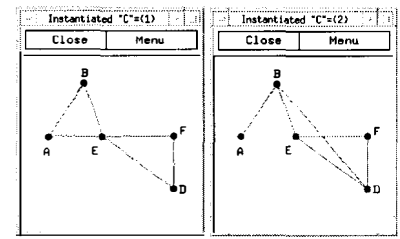

Figure 5: The graphs instantiated by each level of $C$. Since $C$ is conditioned on, this vertex has been eliminated from the graphs.

## 5 CONCLUDING REMARKS

From Figure 1, (1) and (2) it follows that a split graph representation of a split model is not in general unique. A split graph such as in Figure 4 served as a convenient structure in the process of selecting a split model, but it may be hard to understand as visual summary of the model. When it comes to interpretation of a split model we believe that the instantiated graphs such as in Figure 5 are more useful than a split graph itself.

In CSI models and split models as introduced above, all variables are treated on equal footing. As illustrated in Section 4.1 some variables are often purely explanatory while other can be considered as responses. In a more general setting one can have a group of variables which can be considered to be responses to some variables and explanatory to other variables at the same time. Ignoring such distinctions can lead to selecting CSI models which are hard to interpret. Moreover, meaningful and relevant models which are not (log-linear) CSI models, but which nonetheless can be interpreted in terms of context specific independencies can be overlooked.

Boutilier et al. (1996) take a recursive approach to models with context specific independence structures. This means assuming an ordering of the variables such that the joint probability can be given as a product of univariate conditional probabilities. The context specific independencies are entailled locally in the conditional probability tables. This approach has a strong resemblance with classification and regression trees, (Breiman, Friedman, Olshen and Stone 1984).

Their approach on the other hand might be too restrictive. For example, the response variables may relate to each other in a symmetric way such that assuming an ordering of the responses is unreasonable.

Models accounting for the distinction between variables being responses, intermediate and explanatory are discussed by e.g. Wermuth and Lauritzen (1983), Asmussen and Edwards (1983) and Højsgaard and Thiesson (1995). The essential idea is that of conditioning on the explanatory variables in successively larger undirected models and finally combining the results. Split models fit nicely into this framework. With this approach, models with such a recursive structure can be dealt with in Y$\mathcal{GG}$DRASIL.

There is a need for computational architectures which are capable of exploiting the special structures of split models (and CSI models) in connection with propagation. In special cases, for instance when a model admits a complete tree structure with graphs on the leaves of the tree, existing techniques can be applicable. The general case is more delicate, but it can be noted that Eriksen (1999) has made some advances in this connection.

### Acknowledgements

The author is indebted to Jens Henrik Badsberg, the creator of CoCo and Xlisp+CoCo.